\documentclass[runningheads]{llncs}
\usepackage[T1]{fontenc}
\usepackage{graphicx}
\usepackage{booktabs}
\usepackage[misc]{ifsym}
\newcommand{\corr}{(\Letter)}
\usepackage{mwe}
\usepackage{amsmath} 
\usepackage{amssymb}
\usepackage{multicol}
\usepackage{multirow}

\begin{document}

\title{Hierarchical Information-Guided Spatio-Temporal Mamba for Stock Time Series Forecasting}

\titlerunning{Hierarchical Information-Guided Spatio-Temporal Mamba}


\author{Wenbo Yan\inst{1,2} \and
Shurui Wang\inst{1,2} \and
Ying Tan\inst{1,3,4,5} \corr }

\institute{School of Intelligence Science and Technology, Peking University, Beijing
\and
Computational Intelligence Laboratory
\and
Institute for Artificial Intelligence
\and
National Key Laboratory of General Artificial Intelligence
\and
Key Laboratory of Machine Perceptron (MOE)
}

\maketitle              

\begin{abstract}

Mamba has demonstrated excellent performance in various time series forecasting tasks due to its superior selection mechanism. Nevertheless, conventional Mamba-based models encounter significant challenges in accurately predicting stock time series, as they fail to adequately capture both the overarching market dynamics and the intricate interdependencies among individual stocks. To overcome these constraints, we introduce the Hierarchical Information-Guided Spatio-Temporal Mamba (HIGSTM) framework. HIGSTM introduces Index-Guided Frequency Filtering Decomposition to extract commonality and specificity from time series. The model architecture features a meticulously designed hierarchical framework that systematically captures both temporal dynamic patterns and global static relationships within the stock market. Furthermore, we propose an Information-Guided Mamba that integrates macro informations into the sequence selection process, thereby facilitating more market-conscious decision-making. Comprehensive experimental evaluations conducted on the CSI500, CSI800 and CSI1000 datasets demonstrate that HIGSTM achieves state-of-the-art performance.

\end{abstract}

\section{Introduction}
Stock time series forecasting, a pivotal component in investment decision-making, continues to be a primary research focus. The advancement of deep neural networks has driven the development of diverse models to tackle this challenge. Early approaches included RNN-based models such as TPA-LSTM\cite{TPALSTM} and CNN-based models like TCN\cite{tcn}. Later, Transformer-based models, including Crossformer\cite{crossformer}, iTransformer\cite{liu2023itransformer}, and PatchTST\cite{patchtst}, demonstrated robust performance in time series forecasting. With the introduction of Mamba\cite{gu2023mamba}, Mamba-based models such as Bi-Mamba\cite{bimamba} and TimeMachine\cite{ahamed2024timemachine} have emerged as highly effective solutions.

Mamba-based models demonstrate exceptional capabilities in efficient long-sequence modeling. However,  given the distinctive characteristics of the stock market, there exists a pronounced interdependence among stocks within the same market. Modeling each stock's time series independently fails to capture the comprehensive market dynamics. Current Mamba models encounter two primary limitations in stock forecasting: 1. They are unable to model the influence of related stocks on each individual stock during node modeling, and 2. Their selection mechanisms depend exclusively on historical time series, lacking the capacity to integrate additional information to improve selection performance.

To address this issue, we propose the Hierarchical Information-Guided Spatio-Temporal Mamba (HIGSTM). HIGSTM incorporates a hierarchical structure that models the entire stock market from three perspectives: individual, temporal, and global. On one hand, it hierarchically aggregates neighborhood information of stock nodes, capturing both dynamic and static inter-stock relationships. On the other hand, we introduce an information-guided Mamba structure, which progressively extracts time step macro information and global macro information, integrating it into the sequence selection process.

We propose an Index-Guided Frequency Filtering Decomposition that transforms stock time series and indices into the frequency domain. Using stock indices, we derive filter parameters to decompose time series into commonality and specificity components. Each node is initially modeled independently via a Mamba block. Subsequently, the Temporal Information-Guided Spatio-Temporal Mamba (TIGSTM) integrates dynamic temporal information through: (1) leveraging sequence specificity to construct a sparse time-varying relationship graph, enabling dynamic neighborhood aggregation, and (2) aggregating time step macro information from commonality to guide Mamba's sequence selection. The Global Information-Guided Spatio-Temporal Mamba (GIGSTM) incorporates static global information by: (1) aggregating specificity across all time steps to form a global static relationship graph for comprehensive neighborhood aggregation, and (2) consolidating commonality to generate global macro information for enhanced sequence selection guidance. In summary, the main contributions are as follows:

\begin{itemize}
    \item We propose an Index-Guided Frequency Filtering Decomposition method to effectively extract commonality and specificity from time series.
\item We introduce a Hierarchical Information-Guided Spatio-Temporal Mamba structure, which extracts node-related relationships and macro-level information across multiple perspectives, aggregates neighborhood information, and enhances sequence selection.
\item We conduct comprehensive experiments on multiple real-world stock datasets, demonstrating the superior performance of our model.
\end{itemize}

\section{Preliminary}
\subsection{Definition of Stock Spatial-Temporal Forecasting Problem}
\label{dfstp}

We formulate the \textbf{Stock Spatial-Temporal Forecasting Problem} as follows: Given historical data across $T$ time steps for $N$ stocks, where each stock at each time step is described by $F$ features, the dataset is represented as $\textbf{X} \in \mathbb{R}^{N \times T \times F}$. The index time series for market are denoted as $\mathbf{I} \in \mathbb{R}^{T \times 1}$. 

Moreover, stocks demonstrate complex interconnections that collectively constitute a graph structure. In this framework, each stock represents a node, while the correlation between stocks defines the edges. These relationships are captured by the adjacency matrix $\mathbf{\mathcal{G}} \in \mathbb{R}^{N \times N}$, where $\mathbf{\mathcal{G}}_{ij}$ quantifies the correlation between stock $i$ and stock $j$. When an edge connects stock $i$ and stock $j$ ($\mathbf{\mathcal{G}}_{ij} \neq 0$), they are identified as neighboring stocks. The set of neighbors for stock $i$ is formally defined as $\mathbf{U}_i = \{j \ | \ j \neq i \ \text{and} \ \mathbf{\mathcal{G}}_{ij} \neq 0\}$.

\subsection{Definition of Broadcast}

Considering the multi-scale information aggregation presented in the article, various data components may exhibit dimensional inconsistencies. Following the definition in Section \ref{dfstp}, we normalize all variables to conform to the three-dimensional structure [N, T, F]. We introduce matrix broadcasting as the operation of replicating a matrix along absent dimensions to achieve the target [N, T, F] format. This operation is formally expressed as:
\begin{equation}
    \mathbf{\mathcal{B}}^{(Dims)}(*)
\end{equation}

For example, $\mathbf{\mathcal{B}}^{(N)}(\mathbf{Z_{t,f}})$, where $\mathbf{Z_{t,f}} \in \mathbb{R}^{T \times F}$, denotes the operation of adding a dimension to $\mathbf{Z_{t,f}}$ and repeating it $N$ times, resulting in a shape of $\mathbb{R}^{N \times T \times F}$.

\subsection{Fast Fourier Transform and Amplitude Filter}
The Fast Fourier Transform (FFT) is an efficient algorithm for computing the Discrete Fourier Transform (DFT), which maps a time-domain signal to the frequency domain. For ease of reference, we denote the FFT computation process as $FFT(*)$ and its inverse as $iFFT(*)$. Any frequency-domain matrix is denoted as ${\Box}^{f}$, and its corresponding amplitude is denoted as ${\Box}^{f,amp}$.

The amplitude filter selectively enhances or suppresses signals by directly manipulating the amplitude components in the frequency domain. Its inputs are a frequency-domain matrix and filtering parameters. For clarity, we represent the amplitude filtering operation as $\mathbf{\Theta}$.

\subsection{Discretization and State Space Model}\label{ssm}
The computation process of Mamba involves three matrices: the state transition matrix $\mathbf{A}$, the input matrix $\mathbf{B}$, and the output matrix $\mathbf{C}$. Mamba employs the Zero-Order Hold (ZOH) method \cite{gu2023mamba} to convert the continuous parameters $\mathbf{A}$ and $\mathbf{B}$ into discrete parameters $\overline{\mathbf{A}}$ and $\overline{\mathbf{B}}$. We denote this discretization process as:
\begin{equation}
    \begin{aligned}
        \overline{\mathbf{A}}, \overline{\mathbf{B}} &= Discretization(\mathbf{A}, \mathbf{B}, \Delta)\\
        \overline{\mathbf{A}} = \exp {(\Delta\mathbf{A})},& \ 
\overline{\mathbf{B}} = (\exp {(\Delta\mathbf{A})}-\mathbf{E})(\Delta\mathbf{A})^{-1}(\Delta\mathbf{B}   )
    \end{aligned}
\end{equation}
where $\mathbf{E}$ is the identity matrix. Additionally, we denote the computation process of the Structured State Space Model (SSM) as $\mathbf{O} = SSM(\overline{\mathbf{A}}, \overline{\mathbf{B}}, \mathbf{C})(\mathbf{X})$, with its core computation process being:
\begin{equation}
    \begin{aligned}
        \mathbf{h}_t &= \mathbf{\overline{A}} \mathbf{h}_{t-1} + \mathbf{\overline{B}} \mathbf{x}_t \\ 
        \mathbf{o}_t &= \mathbf{C} \mathbf{h}_t
    \end{aligned}
\end{equation}


\section{Method}
\begin{figure}[tbp]
    \centering
    \includegraphics[width=\linewidth]{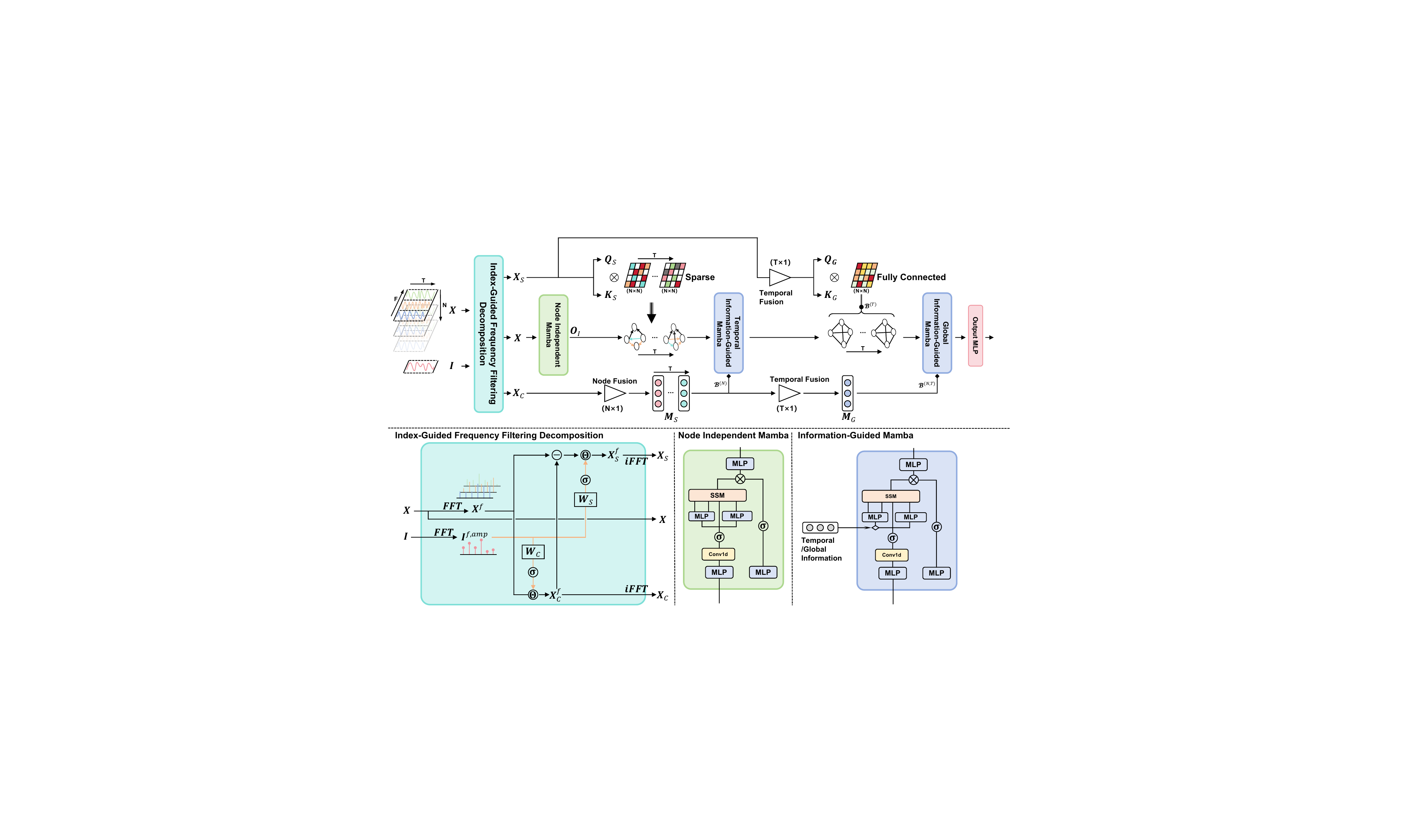}
    \caption{The overview of the proposed Hierarchical Information-Guided Spatio-Temporal Mamba (HIGSTM)}
    \label{fig:stmamba}
\end{figure}

In this section, we introduce our proposed \textbf{Hierarchical Information-Guided Spatio-Temporal Mamba (HIGSTM)}.

\subsection{Index-Guided Frequency Filtering Decomposition}

Stocks within the same market demonstrate both commonality, reflecting shared market trends and macroeconomic information that influences all stocks' future trajectories, and specificity, capturing individual characteristics that are more effective for identifying inter-stock relationships. Eliminating commonality results in sparser stock relationships, preserving only the strongest correlations. 

However, commonality and specificity are intertwined and challenging to decompose directly without a reference. Our analysis reveals that stock indices, serving as indicators of macroeconomic, can effectively guide this decomposition. Leveraging this insight, we introduce the Index-Guided Frequency Filtering Decomposition. This approach begins by transforming both the stock time series $\mathbf{X}$ and the index time series $\mathbf{I}$ into the frequency domain through the Fast Fourier Transform (FFT).

\begin{equation}
    \begin{aligned}
            \mathbf{X}^{f} = FFT(\mathbf{X}),
            &\ \mathbf{I}^{f} = FFT(\mathbf{I}) \\
             \mathbf{I}^{f,amp} &= |\mathbf{I}^{f}|
    \end{aligned}
\end{equation}
where $\mathbf{X}^{f}$ and $\mathbf{I}^{f}$ are the frequency-domain representations of the stock series and the index series, respectively, $\mathbf{I}^{f,amp}$ is the amplitude of the index series.


From a frequency-domain perspective, decomposing commonality and specificity involves analyzing signal intensity across different frequencies. We utilize the index series' amplitude to derive the amplitude filter parameters, which are then applied to filter the stock series.
\begin{equation}
    \mathbf{X}^{f}_{c} = \mathbf{X}^{f} \mathbf{\Theta} \sigma(\mathbf{W}_{c}^\mathsf{T}\mathbf{I}^{f,amp})
\end{equation}
where $\mathbf{W}_{c}$ is a linear transformation matrix, $\sigma$ denotes the sigmoid activation function, $\mathbf{\Theta}$ represents the filtering operation, and $\mathbf{X}^{f}_{c}$ corresponds to the commonality. The filter parameters learned from index series' amplitude ensures that the commonality aligns more closely with macroeconomic market characteristics. Additionally, we focus on isolating the specificity from this decomposition.
\begin{equation}
    \mathbf{X}^{f}_{s} = (\mathbf{X}^{f}-\mathbf{X}^{f}_{c}) \mathbf{\Theta} (1-\sigma(\mathbf{W}_{s}^\mathsf{T}\mathbf{I}^{f,amp}))
\end{equation}
where $\mathbf{W}_{s}$ is a linear mapping matrix and $\mathbf{X}^{f}_{s}$ denotes the specificity. Subtracting $\mathbf{X}^{f}_{c}$ from $\mathbf{X}^{f}$ removes the commonality, preserving each stock's specificity. Leveraging index information, we further eliminate common market trends by deriving an additional amplitude filter. Ultimately, we obtain the decomposed commonality series $\mathbf{X}_{c}$ and specificity series $\mathbf{X}_{s}$ via the inverse Fast Fourier Transform (iFFT).
\begin{equation}
    \begin{aligned}
            \mathbf{X}_{c} = iFFT(\mathbf{X}^{f}_{c}),\ 
            \mathbf{X}_{s} = iFFT(\mathbf{X}^{f}_{s})
    \end{aligned}
\end{equation}

\subsection{Node Independent Mamba}
Following series decomposition, we initially model each stock's time series independently, deliberately excluding inter-stock correlations. This strategy prevents the premature introduction of neighboring node information, which could obscure intrinsic series features and compromise the stock's uniqueness. This methodology mirrors the subjective stock evaluation process: a stock's intrinsic characteristics fundamentally determine its future trajectory, and even in favorable market conditions, a fundamentally weak stock will not yield positive predictions.

We use a Mamba block \cite{gu2023mamba} to independently model each time series, taking the raw stock time series $\mathbf{X}$ as input. $\mathbf{X}$ undergoes a one-dimensional convolution (Conv1d) to extract local features, followed by a Linear Projection that maps it to the input matrix $\mathbf{B}_{In}$, the output matrix $\mathbf{C}_{In}$, and the discretized time step $\Delta_{In}$.
\begin{equation}
    \begin{aligned}
        \mathbf{X}_{In} &= \mu(\text{Conv1d}(\mathbf{X})) \\
        \mathbf{B}_{In} &= \mathbf{W}^\mathsf{T}_{B,I}\mathbf{X}_{In}, \quad \mathbf{C}_{In} = \mathbf{W}^\mathsf{T}_{C,I}\mathbf{X}_{In} \\
        \Delta_{In} &= softplus(\mathbf{W}^\mathsf{T}_{\Delta,I}\mathbf{X}_{In} + \mathbf{A}_{In})
    \end{aligned}
\end{equation}
where $\mathbf{W}_{B,I}$, $\mathbf{W}_{C,I}$, and $\mathbf{W}_{\Delta,I}$ are linear transformation matrices, $\mu(*)$ is the SiLU activation function, $softplus(*)$ denotes the Softplus activation function, and $\mathbf{A}_{In}$ is an optimizable matrix. Subsequently, the matrices $\mathbf{A}_{In}$ and $\mathbf{B}_{In}$ are discretized into $\overline{\mathbf{A}}_{In}$ and $\overline{\mathbf{B}}_{In}$.
\begin{equation}
    \begin{aligned}
\overline{\mathbf{A}}_{In}, \overline{\mathbf{B}}_{In} = Discretization(\mathbf{A}_{In}, \mathbf{B}_{In}, \Delta_{In})
    \end{aligned}
\end{equation}

Then, $\overline{\mathbf{A}}$, $\overline{\mathbf{B}}$, $\mathbf{C}$, and $\mathbf{X}$ are fed into the State Space Model (SSM)\cite{gu2023mamba}.
\begin{equation}
    \mathbf{O}_{In} = SSM(\overline{\mathbf{A}}_{In}, \overline{\mathbf{B}}_{In}, \mathbf{C}_{In})(\mathbf{X}_{In})
\end{equation}
where the $SSM(*)$ process is detailed in Section \ref{ssm}. Within the Node Independent Mamba, we utilize the Mamba block to model each stock node independently, enabling exploration of intrinsic node characteristics while preventing neighboring information from obscuring the node's unique features.

\subsection{Temporal Information-Guided Spatio-temporal Mamba}
The stock market operates as an integrated system, necessitating the consideration of inter-stock interactions. We introduce the Temporal Information-Guided Spatio-temporal Mamba block \textbf{(TIGSTM)}, capturing both inter-stock relationships within each time step and the influence of time step macro information on individual stocks. TIGSTM integrates two key components: Temporal Section Sparse Neighbor Aggregation and the Temporal Information-Guided Selective State Space Model.

\noindent\textbf{Temporal Section Sparse Neighbor Aggregation} 
We utilize the decomposed specificity series $\mathbf{X}_{s}$ to capture inter-stock relationships. Given their dynamic evolution over time, we implement an attention mechanism to establish the correlation at each time step.
\begin{equation}
\begin{aligned}
        \mathbf{Q}_{S}=\mathbf{W}^\mathsf{T}_{Q,S}\mathbf{X}_{s}, \ & \mathbf{K}_{S}=\mathbf{W}^\mathsf{T}_{K,S}\mathbf{X}_{s}\\
    \mathbf{\mathcal{G}}_{S}=softmax(topK(&\frac{\mathbf{Q}_{S}^\mathsf{T}\mathbf{K}_{S}}{\sqrt{d_{kS}}},30\%))+\mathbf{E}_{S}
\end{aligned}
\end{equation}
where $\mathbf{W}_{Q,S}$ and $\mathbf{W}_{K,S}$ denote linear mapping matrices, $\sqrt{d_{kS}}$ represents the dimension of the $\mathbf{K}_{S}$ vector, $\mathbf{E}_{S}$ is the identity matrix, and $topK()$ ensures each node retains only the neighbors with the highest attention weights. The specificity series $\mathbf{X}_{s}$ eliminates commonality, resulting in sparser yet more prominent node relationships. We enforce sparsity in the graph at each time step, restricting each node to retain only 30\% of its neighbors. This enables each stock to aggregate information from its neighbors at each time step.
\begin{equation}
    \mathbf{X}_{S,Agg} = \mathbf{\mathcal{G}}_{S}\mathbf{O}_{In}
\end{equation}

The time series after information aggregation is denoted as $\mathbf{X}_{S,Agg}$.

\noindent\textbf{Temporal Information-Guided Selective State Space Model}
Mamba's Selection mechanism dynamically generates the parameters of the state space model (SSM) based on the time step, with the matrices being derived from the time series, significantly enhancing SSM performance. We introduce the Temporal Information-Guided Selective State Space Model, which integrates the time step macro information into the selection mechanism to guide sequence selection. Initially, we aggregate node information from the commonality series $\mathbf{X}_{c}$ to form the time step macro information.
\begin{equation}
    \mathbf{M}_{S}= \mathbf{W}_{F,S}^\mathsf{T}(\mathbf{W}^\mathsf{T}_{(N\times1)}\mathbf{X}_{c})+\mathbf{b}_{F,S}
\end{equation}
where $\mathbf{W}_{(N\times1)}$ is an $N\times1$ mapping matrix that aggregates information from all stocks, and $\mathbf{W}_{F,S}$ is a linear mapping matrix that further blends the features to form the macro-level information $\mathbf{M}_{S}$ for each time step. We incorporate $\mathbf{M}_{S}$ into the input matrix and output matrix.
\begin{equation}
    \begin{aligned}
        \mathbf{X}_{S} = \mu(\text{Conv1d}&(\mathbf{X}_{S,Agg})) \\
        \mathbf{B}_S = <\mathbf{W}^\mathsf{T}_{B,S}\mathbf{X}_{S}, \mathbf{\mathcal{B}}^{(N)}(\mathbf{M}_{S})>,&\ 
        \mathbf{C}_S = <\mathbf{W}^\mathsf{T}_{C,S}\mathbf{X}_{S}, \mathbf{\mathcal{B}}^{(N)}(\mathbf{M}_{S})> \\
        \Delta_S = softplus(&\mathbf{W}^\mathsf{T}_{\Delta,S}\mathbf{X}_{S} + \mathbf{A}_S)
    \end{aligned}
\end{equation}
where $\mathbf{W}_{B,S}$ and $\mathbf{W}_{C,S}$ denote linear mapping matrices that project the input sequence $\mathbf{X}_{S,Agg}$ into the input and output matrices, $\mathbf{A}_S$ represents a learnable state transition matrix, and $<\dots>$ indicates matrix concatenation. We broadcast and concatenate the time step macro information $\mathbf{M}_{S}$ into input matrix and output matrix. This ensures that the SSM process incorporates time step macro information for input and output decisions. Subsequently, we discretize the state transition matrix $\mathbf{A}_S$ and input matrix $\mathbf{B}_S$ and feed them into the SSM.
\begin{equation}
    \begin{aligned}
        \overline{\mathbf{A}}_S, \overline{\mathbf{B}}_S &= Discretization(\mathbf{A}_S, \mathbf{B}_S, \Delta_S)\\
        \mathbf{O}_S &= SSM(\overline{\mathbf{A}}_S, \overline{\mathbf{B}}_S, \mathbf{C}_S)(\mathbf{X}_S)
    \end{aligned}
\end{equation}
where $\mathbf{O}_S$ represents the final output of TIGSTM. The TIGSTM propagates information within localized neighborhoods, with each time step maintaining unique adjacency relationships to fully capture dynamic inter-stock correlation. Meanwhile, it extracts time step macro information to guide Mamba's sequence selection mechanism. 

\subsection{Global Information-Guided Spatio-temporal Mamba}
Beyond time-evolving dynamic information, static information also influences the change of the stock. To capture this, we introduce the Global Information-Guided Spatio-temporal Mamba Block \textbf{(GIGSTM)}, which extracts both global stock correlation and global macro information.

\noindent\textbf{Global Neighbor Aggregation} 
Global Neighbor Aggregation models the global inter-stock relationships from the specificity series $\mathbf{X}_{s}$ and enables each stock to aggregate information from its global neighbors. Initially, we aggregate the information across all time steps in the specificity series $\mathbf{X}_{s}$ and construct the global stock correlations through an attention mechanism.
\begin{equation}
\begin{aligned}
    \mathbf{X}_{s,G} = &\mathbf{W}^\mathsf{T}_{(T\times1)}\mathbf{X}_{s}\\
        \mathbf{Q}_{G}=\mathbf{W}^\mathsf{T}_{Q,G}\mathbf{X}_{s,G}, \ & \mathbf{K}_{G}=\mathbf{W}^\mathsf{T}_{K,G}\mathbf{X}_{s,G}\\
    \mathbf{\mathcal{G}}_{G}=softmax(&\frac{\mathbf{Q}_{G}^\mathsf{T}\mathbf{K}_{G}}{\sqrt{d_{kG}}}) + \mathbf{E}_G
\end{aligned}
\end{equation}
where $\mathbf{W}_{(T\times1)}$ denotes a $T\times1$ mapping matrix that integrates information from all time steps to derive global specificity $\mathbf{X}_{s,G}$; $\mathbf{W}_{Q,G}$ and $\mathbf{W}_{K,G}$ represent linear mapping matrices; $\sqrt{d_{kG}}$ indicates the dimension of the $\mathbf{K}_{G}$ vector; $\mathbf{E}_G$ is the identity matrix; and $\mathbf{\mathcal{G}}_G$ represents the adjacency matrix of the learned global correlation graph. By aggregating multiple time step information, inter-node correlation becomes more comprehensive. Consequently, we relax the sparsity constraint on $\mathbf{\mathcal{G}}_G$, allowing a fully connected structure. Finally, we broadcast global correlation across all time steps, enabling each stock to aggregate information from all others.
\begin{equation}
    \mathbf{X}_{G,Agg} = \mathbf{\mathcal{B}}^{(T)}(\mathbf{\mathcal{G}}_{G})\mathbf{O}_S
\end{equation}
The time series after information aggregation is denoted as $\mathbf{X}_{G,Agg}$.

\noindent\textbf{Global Information-Guided Selective State Space Model}
Global information plays a pivotal role in time step selection, as identical sequences may yield divergent decisions when incorporating versus excluding market context. We propose the Global Information-Guided Selective State Space Model, which enhances Mamba's selection mechanism through global macro information. Initially, we extract global macro information from each time step's macro data.
\begin{equation}
    \mathbf{M}_{G}= \mathbf{W}_{F,G}^\mathsf{T}(\mathbf{W}^\mathsf{T}_{(T\times1)}\mathbf{M}_{S})+\mathbf{b}_{F,G}
\end{equation}
where $\mathbf{W}_{(T\times1)}$ denotes a $T\times1$ mapping matrix that integrates time step macro information, and $\mathbf{W}_{F,G}$ represents a linear mapping matrix that fuses features to generate global macro information $\mathbf{M}_{G}$. We then incorporate $\mathbf{M}_{G}$ into the input and output matrix to guide the sequence selection process.
\begin{equation}
    \begin{aligned}
        \mathbf{X}_{G} = \mu(\text{Conv1d}&(\mathbf{X}_{G,Agg})) \\
        \mathbf{B}_G = <\mathbf{W}^\mathsf{T}_{B,G}\mathbf{X}_{G}, \mathbf{\mathcal{B}}^{(N,T)}(\mathbf{M}_{G})>,&\ 
        \mathbf{C}_G = <\mathbf{W}^\mathsf{T}_{C,G}\mathbf{X}_{G}, \mathbf{\mathcal{B}}^{(N,T)}(\mathbf{M}_{G})> \\
        \Delta_G = softplus(&\mathbf{W}^\mathsf{T}_{\Delta,G}\mathbf{X}_{G} + \mathbf{A}_G)
    \end{aligned}
\end{equation}
where $\mathbf{W}_{B,G}$ and $\mathbf{W}_{C,G}$ represent linear mapping matrices that project the input sequence $\mathbf{X}_{G,Agg}$ into the input and output matrices, $\mathbf{A}_G$ denotes a learnable state transition matrix, and $<\dots>$ indicates matrix concatenation. We propagate global macro information $\mathbf{M}_{G}$ to each stock's time step and concatenate it with both matrices. This integration ensures the SSM process incorporates global macro information for sequence selection. Finally, we discretize the state transition matrix $\mathbf{A}_G$ and input matrix $\mathbf{B}_G$ for SSM processing.
\begin{equation}
    \begin{aligned}
        \overline{\mathbf{A}}_G, \overline{\mathbf{B}}_G &= Discretization(\mathbf{A}_G, \mathbf{B}_G, \Delta_G)\\
        \mathbf{O}_G &= SSM(\overline{\mathbf{A}}_G, \overline{\mathbf{B}}_G, \mathbf{C}_G)(\mathbf{X}_G)
    \end{aligned}
\end{equation}
where $\mathbf{O}_G$ represents the final output of GIGSTM. The GIGSTM integrates time step macro information to learn both global macro information and static global stock correlations. Through a fully connected graph, each stock accesses information from all others, enhancing its representation. Meanwhile, global macro information guides each stock's time step selection, ensuring predictions incorporate historical patterns while aligning with market trends, thus improving stock modeling.

\subsection{Prediction and Losses}
Unlike conventional sequence prediction methods, we directly map all features output by GIGSTM to the prediction target through a linear layer. Considering the unique characteristics of stock prediction, we employ the mean-deviation prediction approach to achieve better forecasting performance \cite{dpa-stiformer}.
\begin{equation}
    \begin{split}
        &mean = \mathbf{W}_{mean}^\mathsf{T}\mathbf{O}_G+\mathbf{b}_{mean}\\
        &dev = tanh(\mathbf{W}_{dev}^\mathsf{T}\mathbf{O}_G+\mathbf{b}_{dev})\\
        &\hat{\mathbf{Y}} = mean + e^{dev}
    \end{split}
\end{equation}
where $\mathbf{W}_{mean}$ and $\mathbf{W}_{dev}$ are linear mapping matrices, and $\mathbf{b}_{mean}$ and $\mathbf{b}_{dev}$ are the corresponding bias matrices. The prediction target is decomposed into mean and deviation predictions. We use the Pearson correlation coefficient loss $\mathcal{L}_{pearson}$ as the loss function to learn the ranking distribution at each time step.
\begin{equation}
     \mathcal{L}_{pearson}  = -\frac{ (\mathbf{Y} - \bar{\mathbf{Y}})^\mathsf{T} (\hat{\mathbf{Y}} - \bar{\hat{\mathbf{Y}}}) }{ \sqrt{ (\mathbf{Y} - \bar{\mathbf{Y}})^\mathsf{T} (\mathbf{Y} - \bar{\mathbf{Y}}) } \cdot \sqrt{ (\hat{\mathbf{Y}} - \bar{\hat{\mathbf{Y}}})^\mathsf{T} (\hat{\mathbf{Y}} - \bar{\hat{\mathbf{Y}}}) } } 
\end{equation}

\section{Experiments}
\subsection{Experimental Setup}
\subsubsection{Datasets}
We conducted experiments on three datasets:  CSI500, CSI800, CSI1000. Brief statistical information is listed in Table \ref{tab:dataset}. Detailed information about the datasets can be found in the appendix.
\begin{table}[htb]   
    \centering
    \caption{The overall information for datasets}
\resizebox{0.5\linewidth}{!}
{

    \begin{tabular}{c|c|c|c}
        \hline
        $\textbf{Dataset}$ & $\textbf{Samples}$ & $\textbf{Node}$ & \textbf{ Partition}\\
        \hline
        CSI500& 3159 & 500 & 2677/239/243\\ 
        CSI800& 3159 & 800 & 2677/239/243\\ 
        CSI1000& 3159 & 1000 &  2677/239/243\\ 
        \hline

    \end{tabular}
    }

    \label{tab:dataset}

\end{table}
\subsubsection{Baseline}
We compare our model with the following baselines: spatio-temporal models ASTGCN\cite{ASTGCN}, MTGNN\cite{MTGNN}, DCRNN\cite{DCRNN}, STEMGNN\cite{stemGNN}, FC-STGNN\cite{FC-STGNN} and MASTER\cite{li2024master}, and time series models MICN\cite{wang2023micn}, Filternet\cite{filternet}, iTransformer\cite{liu2023itransformer}, TimeMixer\cite{wang2024timemixer}, and TimesNet\cite{wu2022timesnet}. Where MASTER are specialized stock prediction models. Detailed descriptions of these models can be found in the appendix.
\begin{table}[h!]
  \centering

  \caption{Comparison results on CSI500, CSI800 and CSI1000 datasets. $\downarrow$ indicates that the smaller the metric is better. The best result is in bold. The suboptimal results are indicated with an underline.}
    \resizebox{0.8\linewidth}{!}{

    \begin{tabular}{cc|cccccc}
    \toprule

        \multicolumn{2}{c|}{\multirow{2}[4]{*}{}} & \multicolumn{6}{c}{\textbf{CSI500}} \\
\cmidrule{3-8}    \multicolumn{2}{c|}{} & \textbf{IC} & \textbf{PNL} & \textbf{MAXD}$\downarrow$ & \textbf{SHARPE} & \textbf{WINR} & \textbf{PL} \\
    \midrule
    \multicolumn{1}{c}{\multirow{6}[2]{*}{Spatio-Temporal}} 
    & \textbf{ASTGCN} & 0.0434  & 0.0811  & 0.1387  & 0.4676  & 0.5110  & 1.0899  \\
    &\textbf{MTGNN}&0.0503&0.0797&	0.1582&	0.6963&	0.5232&	1.1866	 \\
    &\textbf{DCRNN}&0.0521&0.1401&	0.1537&	0.9265&	0.5162&	1.1688  \\
    &\textbf{STEMGNN} &0.0479&0.0708&	0.1308&	0.7475&	0.4699&	1.0375  \\
          & \textbf{FC-STGNN} & 0.0420  & 0.0946  & 0.1441  & 0.5858  & 0.5124  & 1.1102  \\
          & \textbf{MASTER} & 0.0567  & \underline{0.1766}  & 0.1086  & \underline{1.2904}  & \underline{0.5579}  & \underline{1.2493} \\
    \midrule
    \multicolumn{1}{c}{\multirow{5}[1]{*}{Temporal }} & \textbf{MICN} & \underline{0.0668}  & 0.0834  & 0.1390  & 0.6408  & 0.5165  & 1.1197  \\
          & \textbf{Filternet} & 0.0464  & 0.1345  & \textbf{0.0688 } & 0.8863  & 0.5165  & 1.1688  \\
          & \textbf{iTransformer} & 0.0400  & 0.1214  & 0.1412  & 0.8340  & 0.4835  & 1.1574  \\
          & \textbf{TimeMixer} & 0.0472  & 0.1044  & 0.1436  & 0.7657  & 0.5041  & 1.1398  \\
          & \textbf{TimesNet} & 0.0346  & 0.0516  & 0.1250  & 0.3374  & 0.5041  & 1.0621  \\
          \midrule
          & \textbf{HIGSTM} & \textbf{0.0791 } & \textbf{0.2619 } & \underline{0.1028}  & \textbf{1.4846 } & \textbf{0.5596 } & \textbf{1.3207 } \\
    \midrule

    \multicolumn{2}{c|}{\multirow{2}[4]{*}{}} & \multicolumn{6}{c}{\textbf{CSI800}} \\
\cmidrule{3-8}    \multicolumn{2}{c|}{} & \textbf{IC} & \textbf{PNL} & \textbf{MAXD}$\downarrow$ & \textbf{SHARPE} & \textbf{WINR} & \textbf{PL} \\
    \midrule

\multicolumn{1}{c}{\multirow{6}[2]{*}{Spatio-Temporal}} & \textbf{ASTGCN} & 0.0431&0.0712&0.1380&0.5149&0.5027&1.1140 \\
    &\textbf{MTGNN} &0.0433&0.0939&0.1279&0.6682&0.5091&1.1409 \\
    &\textbf{DCRNN}& 0.0373&0.0482&0.1374&0.3406&0.5098&1.0693 \\
    &\textbf{STEMGNN} &0.0364&0.0594&0.1289&0.3849&0.5032&1.0809 \\
          & \textbf{FC-STGNN} & 0.0285&0.0312&0.1468&0.1803&0.5124&1.0322  \\
          & \textbf{MASTER} & \underline{0.0628}&\underline{0.1603}&\underline{0.1116}&\underline{1.2076}&0.4959&\underline{1.2363}  \\
    \midrule
    \multicolumn{1}{c}{\multirow{5}[1]{*}{Temporal }} & \textbf{MICN}& 0.0380&0.0212&0.1493&0.1614&\underline{0.5165}&1.0288  \\
          & \textbf{Filternet} &  0.0302&0.0951&0.1255&0.6352&0.5083&1.1188 \\
          & \textbf{iTransformer} & 0.0457&0.0314&0.1348&0.2192&0.4752&1.0397 \\
          & \textbf{TimeMixer} & 0.0338&0.0487&0.1280&0.3028&0.5083&1.0548   \\
          & \textbf{TimesNet} & 0.0253&0.0311&0.1427&0.2112&0.4793&1.0387  \\
          \midrule
          & \textbf{HIGSTM} & \textbf{0.0708}&\textbf{0.1918}&\textbf{0.1040}&\textbf{1.2602}&\textbf{0.5263}&\textbf{1.3617} \\
    \midrule
    
    \multicolumn{2}{c|}{\multirow{2}[3]{*}{}} & \multicolumn{6}{c}{\textbf{CSI1000}} \\
\cmidrule{3-8}    \multicolumn{2}{c|}{} & \textbf{IC} & \textbf{PNL} & \textbf{MAXD}$\downarrow$ & \textbf{SHARPE} & \textbf{WINR} & \textbf{PL} \\
    \midrule
    \multicolumn{1}{c}{\multirow{6}[2]{*}{Spatio-Temporal}} & \textbf{ASTGCN} & 0.0694  & 0.1313  & 0.1295  & 0.9434  & 0.4932  & 1.0884  \\
    
        &\textbf{MTGNN}&	0.0701 &0.1108&	0.1406&	1.1415&	0.5098&	1.2393 \\
    &\textbf{DCRNN}&	0.0726&0.1426&	0.1614&	1.1791&	0.5531&	1.2593 \\
    &\textbf{STEMGNN}&	0.0688&0.1392&	0.1021&	1.1334&	0.4867&	1.2084 \\
          & \textbf{FC-STGNN} & 0.0733  & 0.1514  & 0.1365  & 0.9339  & \underline{0.5537}  & 1.1709  \\
          & \textbf{MASTER} &0.0850  & 0.1843  & \underline{0.0995}  & \underline{1.3382}  & 0.5455  & \underline{1.2601} \\
    \midrule
    \multicolumn{1}{c}{\multirow{5}[1]{*}{Temporal }} & \textbf{MICN} & \underline{0.0876}  & 0.1779  & 0.1467  & 1.2339  & 0.5455  & 1.2359  \\
          & \textbf{Filternet} & 0.0723  & \underline{0.1883}  & 0.1312  & 1.2715  & 0.5413  & 1.2392  \\
          & \textbf{iTransformer} & 0.0747  & 0.1642  & 0.1407  & 1.1638  & 0.5289  & 1.2226  \\
          & \textbf{TimeMixer} & 0.0797  & 0.1580  & 0.1178  & 1.0681  & 0.5289  & 1.2007  \\
          & \textbf{TimesNet} & 0.0766  & 0.1512  & 0.1092  & 1.0244  & 0.5372  & 1.1922  \\
    \midrule
          & \textbf{HIGSTM} & \textbf{0.0918 } & \textbf{0.2178 } & \textbf{0.0759 } & \textbf{1.4173 } & \textbf{0.5845 } & \textbf{1.3566 } \\
    \bottomrule      
    \end{tabular}%
}

  \label{tab:main_result}%
\end{table}%

\subsubsection{Metrics}
We evaluate the performances of all baseline by nine metrics which are commonly used in stock prediction task including Information Coefficient (IC), Profit and Loss (PNL), Max Drawdown (MAXD), Sharpe Ratio (SHARPE), Win Rate (WINR), Profit/Loss Ratio (PL).

\subsection{Main Results}

We compare the proposed HIGSTM model with eight baselines across three datasets, and the experimental results are shown in Table 2. Overall, our method achieves the best performance in terms of IC, PNL, SHARPE, WINR, and PL on all datasets. For MAXD, HIGSTM just ranks second on the CSI500 dataset, slightly behind Filternet. On average, IC improves by 11\%, SHARPE by 10\%, and PNL and PL by over 6\%. Additionally, the model’s trading win rate increases by at least 2\%, which represents a substantial improvement for stock trading.

More specifically, HIGSTM significantly outperforms other models in IC, with improvements of up to 18\%, indicating that our method provides superior predictions. Compared to temporal models, HIGSTM incorporates cross-sectional and global information to guide sequence modeling, allowing each sequence to consider both historical patterns and overall market trends. Compared to spatio-temporal models, our proposed decomposition method extracts sequence specificity, and the hierarchical structure progressively models both dynamic and static relationships among stocks. The improvements in WINR and PL demonstrate that our model has stronger predictive capabilities for top-performing stocks, enabling more accurate identification of high-potential stocks. Regarding MAXD, HIGSTM underperforms Filternet on the CSI500 dataset, suggesting that while HIGSTM delivers higher returns, it also introduces slightly higher risk. However, this is not necessarily a drawback, as evidenced by the SHARPE metric, which measures the ratio of return to risk. HIGSTM significantly outperforms other models in SHARPE, with improvements exceeding 15\%, indicating that the increase in returns far outweighs the associated risk.

\subsection{Ablation Study}

To validate the effectiveness of our model, we conducted ablation study on the CSI500 and CSI1000 datasets, using IC (to measure prediction accuracy) and SHARPE (to measure the return-to-risk ratio) as evaluation metrics. Specifically, we compared the following variants:
\begin{itemize}
    \item \textbf{w/o TIGSTM}: remove Temporal Information-Guided Spatio-temporal Mamba block
    \item \textbf{w/o GIGSTM}: remove Global Information-Guided Spatio-temporal Mamba block
    \item \textbf{w/o T\&GIGSTM}: remove Temporal and Global Information-Guided Spatio-temporal Mamba block
    \item \textbf{w/o Decomposition}: remove the Index-Guided Frequency Filtering Decomposition; both the relationships and macro-level information are extracted directly from the stock time series.
    \item \textbf{w/o Index}: when decomposing the stock time series, the decomposition is not guided by the stock indices.
\end{itemize}
The experimental results are shown in Table 3. It can be observed that removing either TIGSTM or GIGSTM leads to a significant decline in both IC and SHARPE, with similar magnitudes of degradation. This indicates that both temporal cross-sectional information and global information are crucial for prediction. When both TIGSTM and GIGSTM are removed, the model essentially loses its predictive capability, demonstrating the effectiveness of our proposed TIGSTM and GIGSTM. Additionally, removing the Decomposition process results in a 30\% performance drop, highlighting the necessity of extracting commonality and specificity from the time series. Commonality enables better extraction of macro-level information, while specificity helps uncover more prominent relationships. Furthermore, when the Decomposition process is not guided by the Index, the model's performance also declines by over 30\%, underscoring the critical role of the Index in guiding the decomposition. Using the Index to learn filter parameters is an indispensable step.

\begin{table}[htbp]
  \centering
      \caption{Ablation Study}
  \resizebox{0.7\linewidth}{!}{
    \begin{tabular}{c|cc|cc}
    \toprule
          & \multicolumn{2}{c|}{\textbf{CSI500}} & \multicolumn{2}{c}{\textbf{CSI1000}} \\
\cmidrule{2-5}          & \textbf{IC} & \textbf{SHARPE} & \textbf{IC} & \textbf{SHARPE} \\
    \midrule
    \textbf{w/o TIGSTM} & 0.0733  & 0.8262  & 0.0745  & 0.9026  \\
    \textbf{w/o GIGSTM} & 0.0604  & 0.6124  & 0.0701  & 1.0075  \\
    \textbf{w/o T\&GIGSTM} & 0.0409  & 0.0333  & 0.0630  & 0.9172  \\
    \textbf{w/o Decomposition} & 0.0539  & 0.6822  & 0.0594  & 0.4605  \\
    \textbf{w/o Index} & 0.0451  & 0.5711  & 0.0592  & 0.7073  \\
    \midrule
    \textbf{HIGSTM} & \textbf{0.0791 } & \textbf{1.4846 } & \textbf{0.0918 } & \textbf{1.4173 } \\
    \bottomrule
    \end{tabular}%
    }
  \label{tab:ablation}%

\end{table}%

\begin{figure}[!t]
    \centering
    \includegraphics[width=0.7\linewidth]{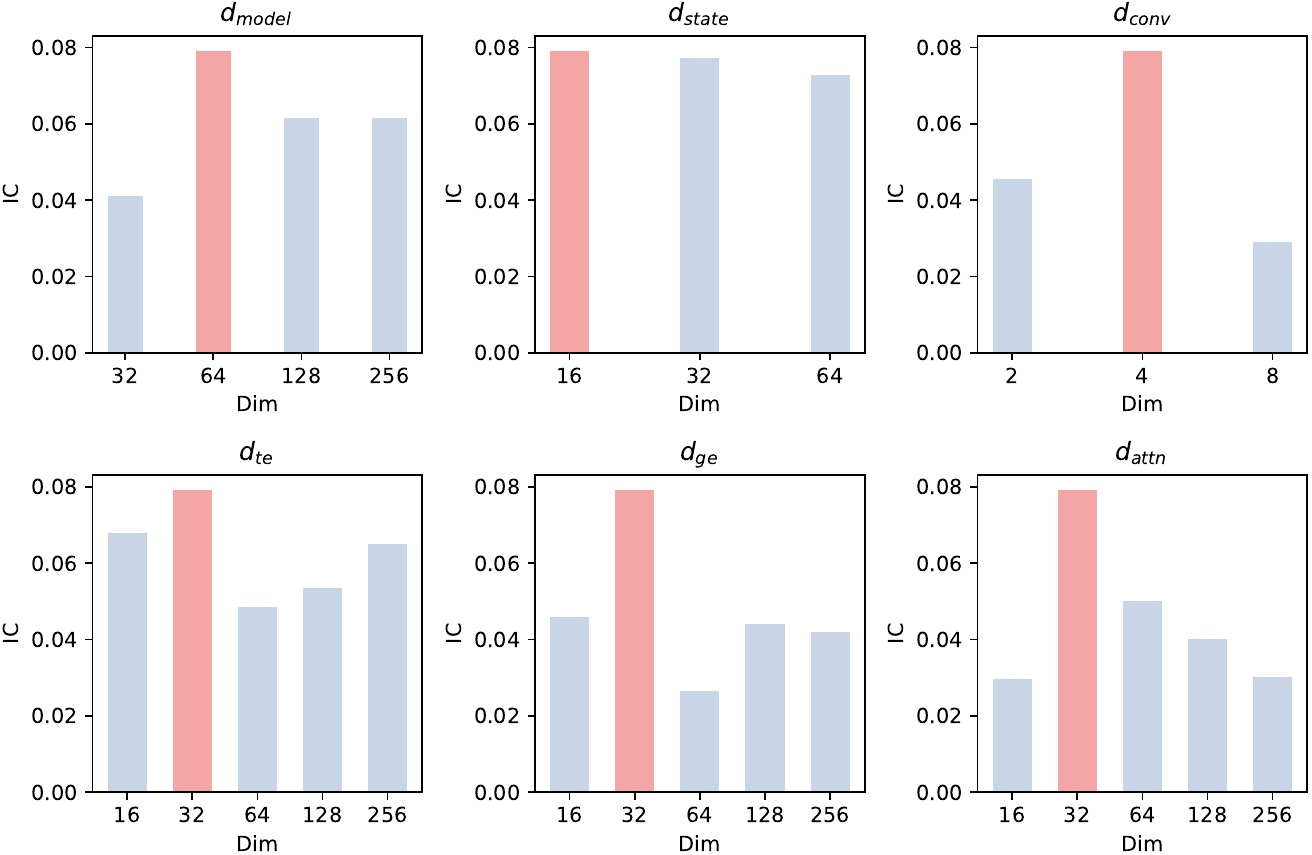}
    \caption{Hyper-parameter Study}
    \label{fig:param}
\end{figure}
\subsection{Hyper-parameter Study}
We conducted a detailed parameter study on the model on CSI500 dataset, and the experimental results are shown in Figure \ref{fig:param}. We investigated the following parameters in HIGSTM: the hidden layer dimension \( d_{model} \), the state dimension \( d_{state} \), the convolution kernel dimension \( d_{conv} \), the macro-level information dimension for each time step \( d_{te} \), the global macro-level information dimension \( d_{ge} \), and the correlation attention dimension \( d_{attn} \). 

The results show that the optimal value for \( d_{model} \) is 64. When \( d_{model} \) is smaller than 64, the dimension is insufficient to capture all features, while when it exceeds 64, the model tends to overfit, leading to a decline in performance. For \( d_{state} \), although the performance differences are minor, a value of 16 is recommended considering computational efficiency. The optimal value for \( d_{conv} \) is 4, as larger convolution kernels can obscure local features, degrading performance. For \( d_{te} \), \( d_{ge} \), and \( d_{attn} \), the optimal dimension is 32. When these parameters are smaller than 32, the model cannot fully capture the required information, while values larger than 32 lead to overfitting, where the guiding information overshadows the unique characteristics of each node.

\subsection{Analysis}

\begin{figure}
    \centering
    \includegraphics[width=\linewidth]{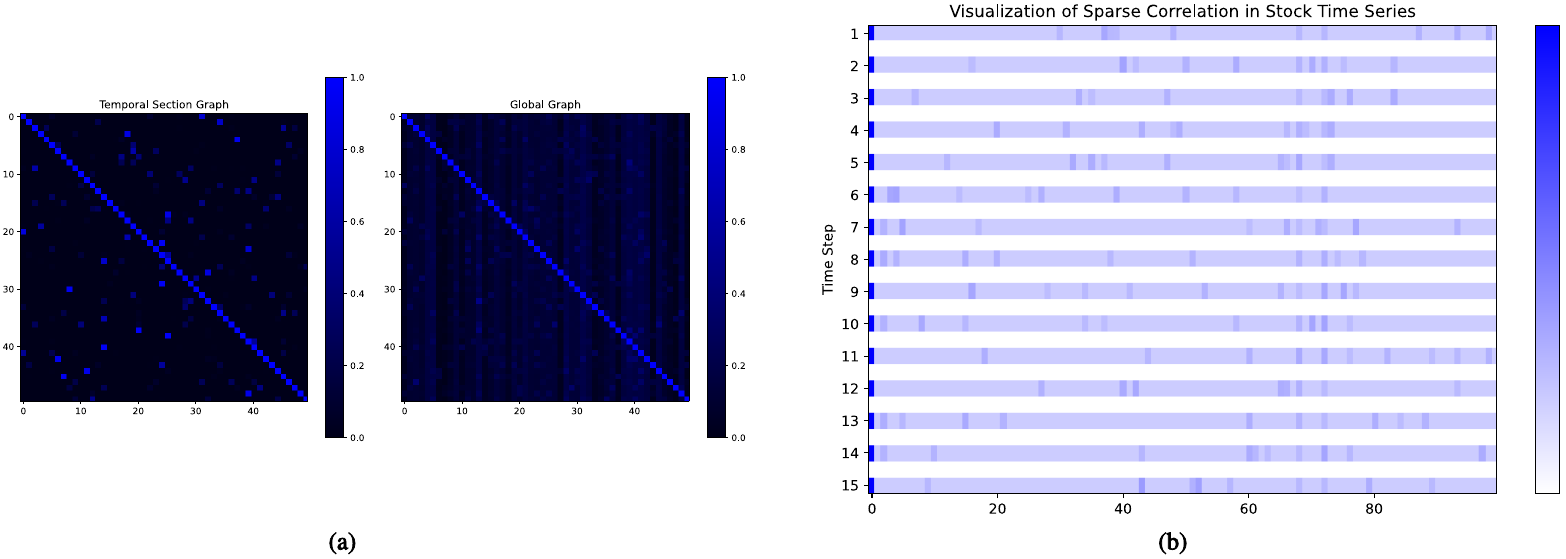}
    \caption{(a)Visualization of Temporal Section and Global Graphs. (b)Visualization of Sparse Correlation in Stock Time Series}
    \label{fig:1stockandtggrph}
\end{figure}

\noindent\textbf{Analysis of Characteristics of Temporal Section and Global Graphs}
We sampled 50 nodes and analyzed the characteristics of both the temporal section and the global graph through visualization. The visualized structure is shown in Figure \ref{fig:1stockandtggrph}(a). It can be observed that the graph learned in each temporal section is sparse and prominent. Each node has only a few neighboring nodes, and most nodes exhibit 1-2 highly correlated neighbors that stand out significantly. This aligns with the model's design intention, which is to identify a small number of truly significant neighbors at each time point, preventing the information from being overshadowed by a large number of neighbors. On the other hand, the Global Graph identifies general static relationships, measuring correlations with all nodes, and does not exhibit significantly correlated neighbors.

\noindent\textbf{Analysis of Dynamic Changes in Correlations}
We selected a stock (stock 0) and visualized its correlations with 100 stocks over time, as shown in Figure \ref{fig:1stockandtggrph}(b). It can be observed that, on one hand, the adjacency relationships of stock 0 change significantly over time steps, indicating that the temporal section graph can capture rapidly changing relationships based on different time steps. On the other hand, stock 0 shares the same neighbors across time steps, demonstrating that the temporal graph is not only capable of capturing fast-changing relationships at the time-step level, but also relationships that persist over a period of time.


\section{Related Work}
\subsection{Temporal Models}
Research in time series forecasting has focused on improving prediction accuracy through various models. Transformer-based models include 
Informer\cite{zhou2021informer}, 
and FEDformer\cite{zhou2022fedformer}, which use sparse attention and decomposition techniques. CNN-based models like TimesNet\cite{wu2022timesnet} extract patterns via frequency and time domain segmentation. MLP-based models such as 
TimeMixer\cite{wang2024timemixer} leverage basis approximation and multi-scale decomposition. Mamba-based models like TimeMachine\cite{ahamed2024timemachine} and Bi-Mamba\cite{bimamba} refine content selection. Other models include PatchTST\cite{patchtst}, which segments time series into patches, and iTransformer\cite{liu2023itransformer}, which redefines time embeddings.

\subsection{Spatial-Temporal Models}
Spatial-temporal forecasting integrates temporal and spatial information for better predictions. 
Convolution-based models such as 
MTGNN\cite{MTGNN}, StemGNN\cite{stemGNN}, 
use gating mechanisms for feature extraction. Complex architectures like ST-GDN\cite{stgdn} improve performance through advanced designs. RNN+GNN models like 
DCRNN\cite{DCRNN} 
combine temporal and spatial modeling. Transformer-based models such as ASTGCN\cite{ASTGCN} 
and STTN\cite{STTN} integrate attention mechanisms for capturing dependencies.

\subsection{Stock Price Forecasting Models}
Stock price forecasting has evolved from traditional methods like ANN\cite{ann} 
and regularization to deep learning. Models like MTDNN\cite{mtdnn} have shown potential. Spatial-temporal frameworks like 
STHAN-SR\cite{sthansr} 
and MASTER\cite{li2024master} capture temporal and variable dependencies.

\section{Conclusion}
In this paper, we propose the Hierarchical Information-Guided Spatio-Temporal Mamba (HIGSTM) to address the limitations of Mamba in stock time series forecasting. HIGSTM decomposes the series into commonality and specificity through Index-Guided Frequency Filtering Decomposition, extracting macro-level information and inter-stock relationships separately. Then, it introduces a hierarchical structure: Node Independent, Temporal Information-Guided Spatio-temporal Mamba, and Global Information-Guided Spatio-temporal Mamba, progressively modeling node features, temporal dynamic correlations, and static global relationships. Our proposed Information-Guided Mamba incorporates time step macro-level information and global information to guide the sequence selection process, ensuring decisions align with the overall market. Experimental results on real-world stock market data fully demonstrate the effectiveness of our approach.

\bibliographystyle{splncs04}
\bibliography{ref}

\clearpage
\appendix
\section{Basic Informations}
\subsection{Datasets}
\begin{itemize}
    \item CSI500:  CSI500 is a stock dataset that contains the performance of 500 small and medium-sized companies listed on the Shanghai and Shenzhen stock exchanges. It contains daily frequency data for 500 stocks, with a total time step of 3159 days and a feature number of 45. (2010.1.1-2023.12.31)
    \item CSI8000: CSI800 contains daily frequency data for 800 stocks, with a total time step of 3159 and a feature number of 45. (2010.1.1-2023.12.31)
    \item CSI1000: CSI1000 contains daily frequency data for 1000 stocks, with a total time step of 3159 and a feature number of 45. (2010.1.1-2023.12.31)
\end{itemize}
The stock lists are the constituent stocks of the CSI500 Index, CSI300 Index + CSI500 Index and the CSI 1000 Index as of December 31, 2023. The 45 data features are specifically as follows: 
\begin{table}[h]
    \centering
        \caption{Data Features}
    \begin{tabular}{c|c|c|c}
    \toprule
         open& high& low& close\\
         pre\_close& change& pct\_chg& vol  \\
        amount& turnover\_rate& turnover\_rate\_f& volume\_ratio\\
        pe&pe\_ttm& pb& ps\\
        ps\_ttm& dv\_ratio& dv\_ttm& total\_share\\
        float\_share& free\_share& total\_mv& circ\_mv\\
        buy\_sm\_vol&buy\_sm\_amount& sell\_sm\_vol& sell\_sm\_amount\\
        buy\_md\_vol&buy\_md\_amount& sell\_md\_vol& sell\_md\_amount\\
        buy\_lg\_vol&buy\_lg\_amount& sell\_lg\_vol& sell\_lg\_amount\\
        buy\_elg\_vol&buy\_elg\_amount& sell\_elg\_vol& sell\_elg\_amount\\
        net\_mf\_vol&net\_mf\_amount& up\_limit& down\_limit\\
        ma5& ma10& ma15&  ma20\\
        ma25&        industry& & \\
    \bottomrule
    \end{tabular}

    \label{tab:my_label}
\end{table}

\subsection{Baselines}
\begin{itemize}
\item \textbf{ASTGCN}\cite{ASTGCN}: ASTGCN incorporates attention mechanisms and graph convolutional operations to effectively capture both spatial and temporal dependencies in time series data. The attention mechanism helps identify and emphasize the most relevant spatial and temporal features, enhancing the model's ability to forecast complex patterns.

\item \textbf{MTGNN}\cite{MTGNN}: MTGNN leverages graph neural networks and multiple timescales temporal module in its architecture to capture short-term and long-term dynamics simultaneously. 

\item \textbf{DCRNN}\cite{DCRNN}: DCRNN (Diffusion Convolutional Recurrent Neural Network) combines diffusion graph convolution and recurrent neural networks to effectively model spatial and temporal dependencies in spatiotemporal data. It captures spatial dependencies through diffusion convolution on graphs and temporal dependencies using recurrent units like GRU.

\item \textbf{STEMGNN}\cite{stemGNN}: STEMGNN utilizes Graph Fourier Transform to transform the spatial dimension from the spatial domain to the frequency domain and employs Spectral Sequential Cell and Spectral Graph Convolution to capture temporal and spatial features.

\item \textbf{FC-STGNN}\cite{FC-STGNN}: FC-STGNN constructs a decay graph that links variables across all timestamps based on their temporal distances. By employing graph convolution, the model captures intricate temporal and spatial dependencies, making it particularly effective for tasks involving multi-variable time series forecasting.

\item \textbf{MASTER}\cite{li2024master}: MASTER is designed to model multi-scale spatio-temporal dependencies in time series data. It leverages attention mechanisms to dynamically weigh the importance of different scales and features, enabling more accurate and robust predictions.

\item \textbf{MICN}\cite{wang2023micn}: MICN focuses on capturing multi-scale temporal patterns in time series data. It employs interactive convolutional layers to model dependencies across different time scales, improving the model's ability to handle complex and long-range temporal relationships.

\item \textbf{FilterNet}\cite{filternet}: FilterNet utilizes learnable frequency filters to extract temporal patterns from time series data. By selectively allowing certain frequency components to pass through or be attenuated, the model can effectively capture and emphasize the most relevant temporal features, leading to improved forecasting accuracy.

\item \textbf{iTransformer}\cite{liu2023itransformer}: iTransformer incorporates interactive self-attention mechanisms to model the interactions between different time series. This approach allows the model to better capture the mutual influences among various time series, enhancing its ability to forecast complex multi-variable time series data.

\item \textbf{TimeMixer}\cite{wang2024timemixer}: TimeMixer is a fully MLP-based architecture designed for time series forecasting. It decomposes time series into multiple scales and independently processes the seasonal and trend components. By mixing these components at different scales, the model achieves a more comprehensive understanding of the underlying temporal patterns.

\item \textbf{TimesNet}\cite{wu2022timesnet}: TimesNet leverages multi-scale temporal representations to capture patterns at different time scales. It employs a hierarchical structure to handle both short-term and long-term dependencies effectively, making it particularly well-suited for tasks requiring robust and scalable time series forecasting.
\end{itemize}
\subsection{Metrics}
    \noindent \textbf{IC}: the Pearson correlation between predicted ranking scores and real ranking scores, widely used to evaluate the performance of stock ranking.
    $$IC = \frac{{\text{Cov}(\hat{y},y)}}{{\sigma_{\hat{y}} \sigma_y}}$$
    
    \noindent \textbf{PNL}: the aggregate profits and losses of trading strategies. $r$ represents the return rate of each transaction and $N_d$ represents the number of transactions.
    $$PNL= \sum_{n=1}^{N_d}{r_n}$$
    
    
    
    \noindent \textbf{MAXD}: the largest decline in portfolios from peak to trough.  
    $$MAXD= {r_{max}}-{r_{min}} \times 100\%$$
    
    \noindent \textbf{SHARPE}: the ratio of the average and standard deviation of portfolio returns, considering both of profitability and investment risk.
    $$SHARPE = \frac{\mu(r)}{\sigma(r)} \times \sqrt{240}$$ 
    
    \noindent \textbf{WINR}: the ratio of the number of positive purchases to the total number of purchases. 
    $$WINR= \frac{N_{+}}{N_{d}}\times 100\%$$
    where $N_{+}$ represents the number of profitable transactions, and $N_{d}$ represents the total number of transactions.
    
    \noindent \textbf{PL}: the profit/loss ratio is the average profit on winning trades divided by the average loss on losing trades over a specified time period.
    $$PL = \frac{\sum{r_{+}}}{N_{+}} \bigg/ \frac{\sum{r_{-}}}{N_{-}} $$
    where $N_{+}$ represents the number of profitable transactions, and $N_{-}$ represents the number of unprofitable transactions
\subsection{Experimental Setup and Parameter Settings}
Our experimental environment is as follows: Ubuntu 22.04, GPU NVIDIA GeForce RTX 3090 24G, CPU AMD EPYC7282, and 256GB of RAM. We conducted 5 repeated experiments.

We set the parameters of HIGSTM as follows: the hidden layer dimension is $d_{model}=64$, the state dimension is $d_{model}=16$, the convolution kernel dimension is $d_{conv}=4$, the macro-level information dimension for each time step is $d_{te}=32$, the global macro-level information dimension is $d_{ge}=32$, and the correlation attention dimension is $d_{attn}=32$. The input time series length is $l_{in}=16$.

\subsection{Trading Strategies}
Our simulated trading strategy involves selecting the top 10\% of stocks each day, holding them for 10 days, and calculating the returns using equal-weighted simple interest for all stocks, while also taking into account transaction fees.
\section{Case Study}
\begin{figure}
    \centering
    \includegraphics[width=\linewidth]{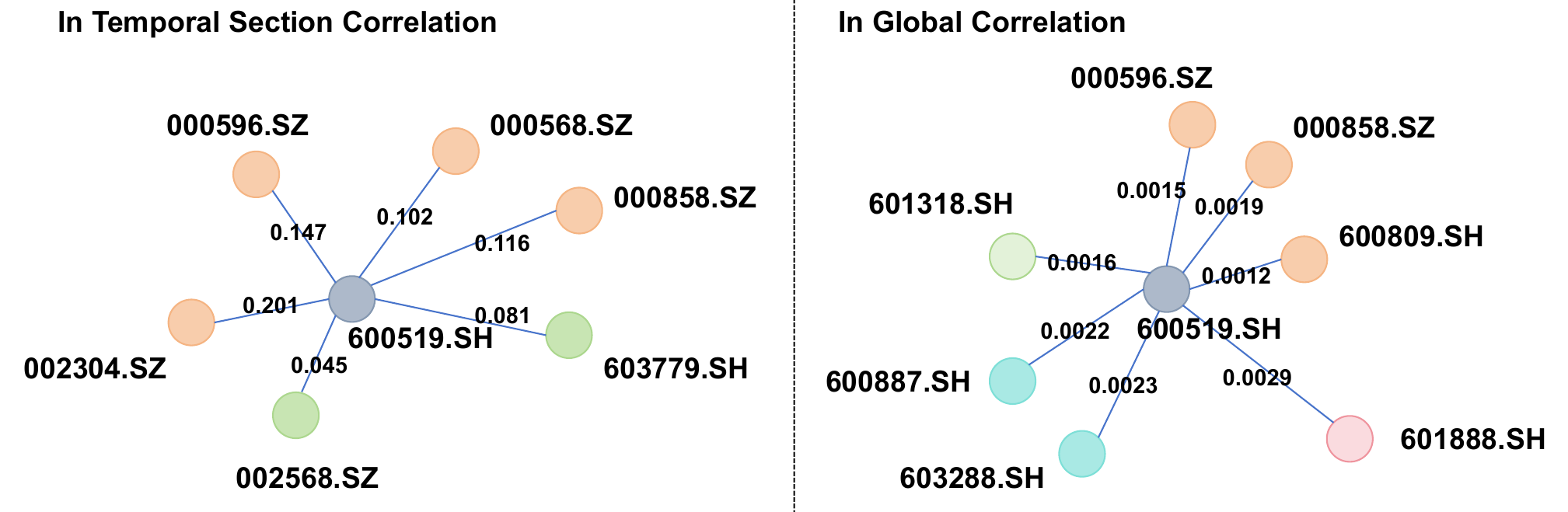}
    \caption{Case: The neighbors of stock 600519.SH learned in the temporal section and the global graph.}
    \label{fig:case1}
\end{figure}
We present the stocks most correlated with 600519.SH in the temporal section and the global graph, as shown in Figure \ref{fig:case1}. The numerical codes in the figure correspond to individual stocks in the Chinese stock market. It can be observed that in the sparse temporal section correlation, the diversity of neighbors is low, but the correlations are prominent. The learned neighbors, such as 000596.SZ (Orange node), are stocks from the same liquor industry as 600519.SH, exhibiting significant correlations. In contrast, the Global Correlation shows higher diversity among neighbors, with no particularly dominant adjacency relationships. For instance, 000596.SZ (Orange node) has a high correlation due to belonging to the same industry as the benchmark stock, 603288.SH (Blue node) is highly correlated as it is also in the beverage sector, 601888.SH (Pink node) is correlated due to its involvement in downstream retail, and 601318.SH (Green node) is correlated as it is another industry leader.

\section{Visualization and Comparison}
\textbf{Compare Graphs Learned using Specificity or Not}\\
\begin{figure}
    \centering
    \includegraphics[width=\linewidth]{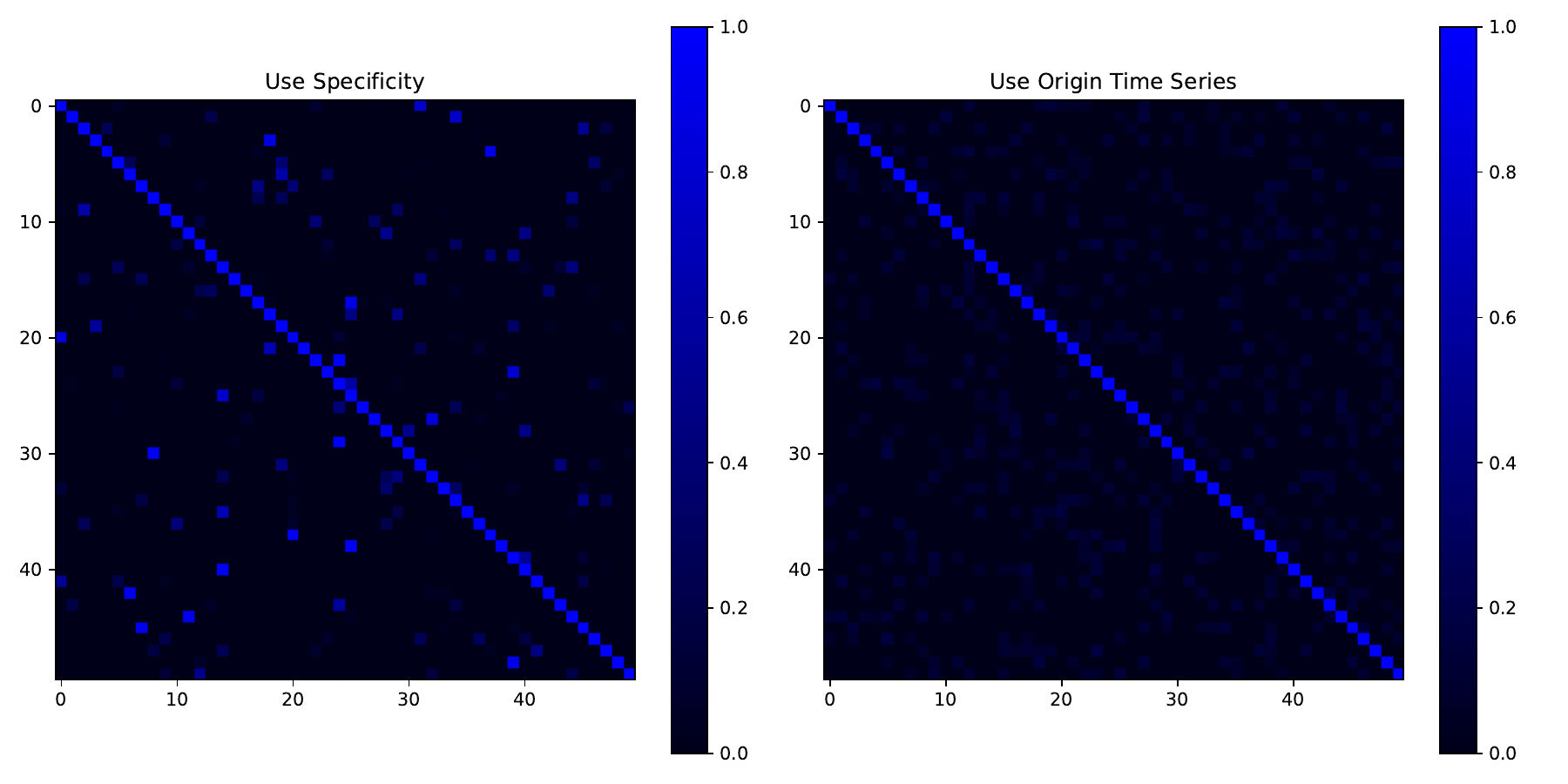}
    \caption{Temporal Section Graphs Learned using Specificity or Origin Time Series}
    \label{fig:uornotu}
\end{figure}
We compared the Temporal Section Graphs learned using either the decomposed Specificity or the Origin Time Series through an attention mechanism. As shown in Figure \ref{fig:uornotu}, when using Specificity, nodes can capture prominent neighbors with high correlations, whereas using the original time series results in no significant differences among neighboring nodes, with no prominent neighbors. This is because the Origin Time Series is influenced by the same market, containing commonality, which obscures the specific correlations. In contrast, Specificity removes the market commonality, retaining only the prominent and significant correlations between features.
\section{Simulated Backtesting Results}
\subsection{Daily Return Rate Analysis}
\begin{figure}[ht]
    \centering
    \includegraphics[width=0.7\linewidth]{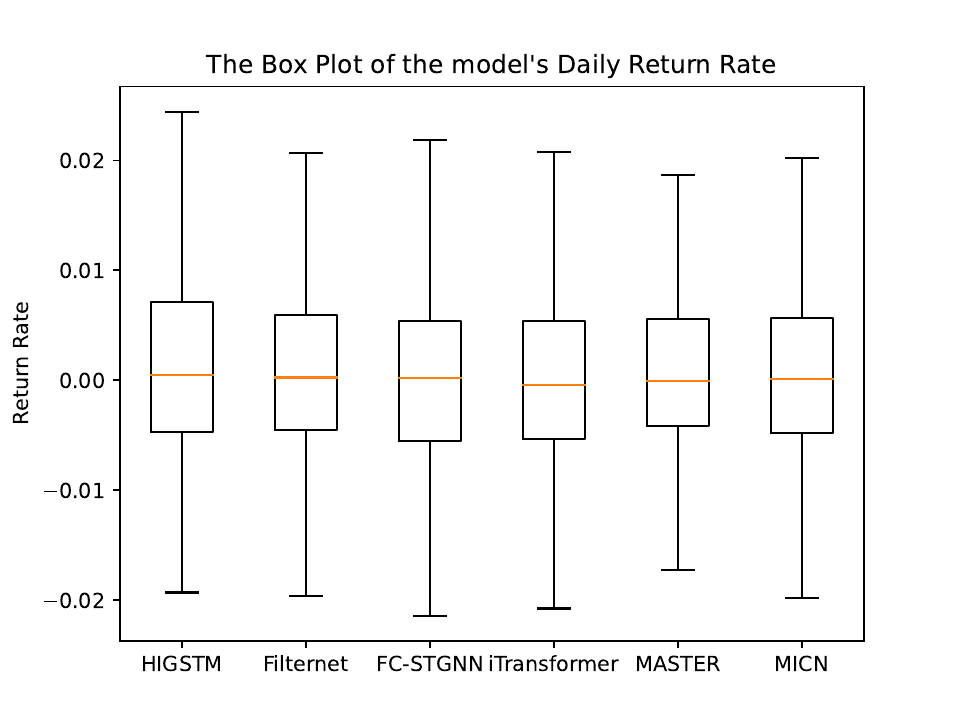}
    \caption{The Box Plot of the model's Daily Return Rate}
    \label{fig:box_plot}
\end{figure}
We plotted the box plots of the daily returns for each model over a year, as shown in Figure \ref{fig:box_plot}. It can be observed that our method, HIGSTM, achieves a significantly higher maximum daily return compared to other baselines, while its minimum return is also higher than that of all models except MASTER. In terms of distribution, the return volatility of our model is slightly greater than that of other models, indicating that HIGSTM carries slightly higher risk, which is consistent with the earlier experimental results. MASTER exhibits a distinct style compared to other models, being a low-risk but low-return model. Therefore, although the minimum return of HIGSTM is lower than that of MASTER, the maximum return of MASTER is significantly lower than that of HIGSTM.
\subsection{One-Year Backtest Return Curve}
\begin{figure}[!h]
    \centering
    \includegraphics[width=\linewidth]{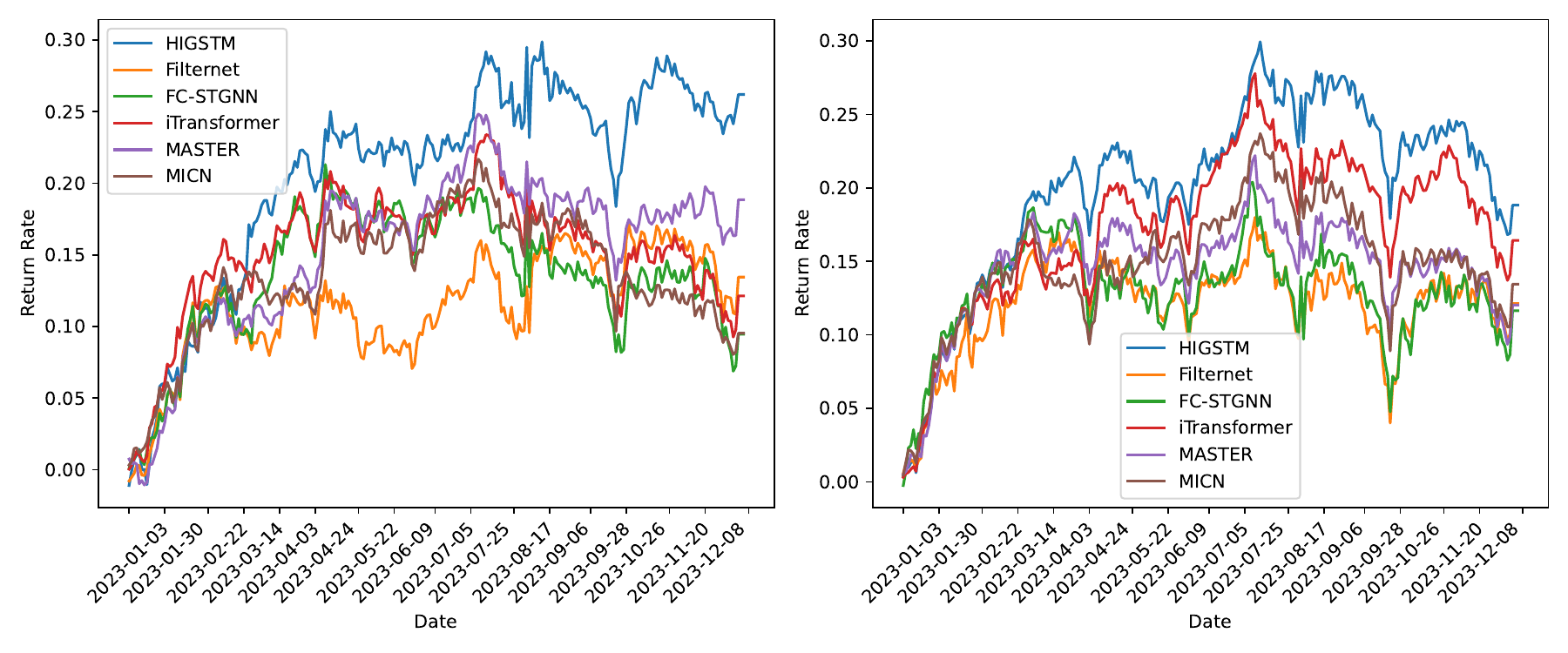}
    \caption{One-Year Backtest Return Curve}
    \label{fig:return_curve}
\end{figure}
We plotted the one-year backtest return curves for the CSI500 and CSI1000 datasets, as shown in Figure \ref{fig:return_curve}. It can be observed that our method achieves significantly higher returns on both the CSI500 and CSI1000 datasets compared to other models. In many upward trends, our model shows notably greater gains than others, such as during the period from September 2023 to October 2023. Additionally, our model successfully avoids drawdowns in many minor correction phases, such as from February 2023 to March 2023. Furthermore, in some significant drawdown periods, our model experiences smaller declines compared to others, such as during the July 2023 to August 2023 interval on the CSI500 dataset.
\end{document}